\documentclass[10pt,twocolumn,letterpaper]{article}

\pdfoutput=1
\usepackage{cvpr}
\usepackage{times}
\usepackage{epsfig}
\usepackage{graphicx}
\usepackage{amsmath}
\usepackage{amssymb}

\newcommand{\xz}{\textcolor{red}}
\newcommand{\xzb}{\textcolor{blue}}
\newcommand{\xzg}{\textcolor{green}}

\usepackage{lineno}
\usepackage{moreverb}
\usepackage{fancyhdr}
\usepackage{calc}
\usepackage{float}
\usepackage{subfigure}
\usepackage{color}
\usepackage[table]{xcolor}
\usepackage{arydshln}
\usepackage{ulem} 

\usepackage{epstopdf}

\usepackage{bm}
\usepackage{setspace}
\usepackage{multirow}

\usepackage[ruled,vlined,linesnumbered,resetcount]{algorithm2e}
\modulolinenumbers[5]

\usepackage{array}
\newcommand{\PreserveBackslash}[1]{\let\temp=\\#1\let\\=\temp}
\newcolumntype{C}[1]{>{\PreserveBackslash\centering}p{#1}}
\newcolumntype{R}[1]{>{\PreserveBackslash\raggedleft}p{#1}}
\newcolumntype{L}[1]{>{\PreserveBackslash\raggedright}p{#1}}

\usepackage[breaklinks=true,bookmarks=false]{hyperref}

\cvprfinalcopy 

\def\cvprPaperID{****} 

\begin{document}

\title{VIFB: A Visible and Infrared Image Fusion Benchmark}

\author{Xingchen Zhang$^{1,2}$, Ping Ye$^1$, Gang Xiao$^1$\\	
	$^1$School of Aeronautics and Astronautics, Shanghai Jiao Tong University\\
	$^2$Department of Electrical and Electronic Engineering, Imperial College London\\
	{\tt\small xingchen.zhang@imperial.ac.uk, \{yeping2018, xiaogang\}@sjtu.edu.cn}
}

\maketitle
\pagestyle{empty}  
\thispagestyle{empty} 

\begin{abstract}
	Visible and infrared image fusion is an important area in image processing due to its numerous applications.~While much progress has been made in recent years with efforts on developing image fusion algorithms, there is a lack of code library and benchmark which can gauge the state-of-the-art.~In this paper, after briefly reviewing recent advances of visible and infrared image fusion,  we present a visible and infrared image fusion benchmark (VIFB) which consists of 21 image pairs, a code library of 20 fusion algorithms and 13 evaluation metrics.~We also carry out extensive experiments within the benchmark to understand the performance of these algorithms.~By analyzing qualitative and quantitative results, we identify effective algorithms for robust image fusion and give some observations on the status and future prospects of this field.
\end{abstract}

\section{Introduction}
\begin{figure}
	\centering
	\includegraphics[width=7.5cm]{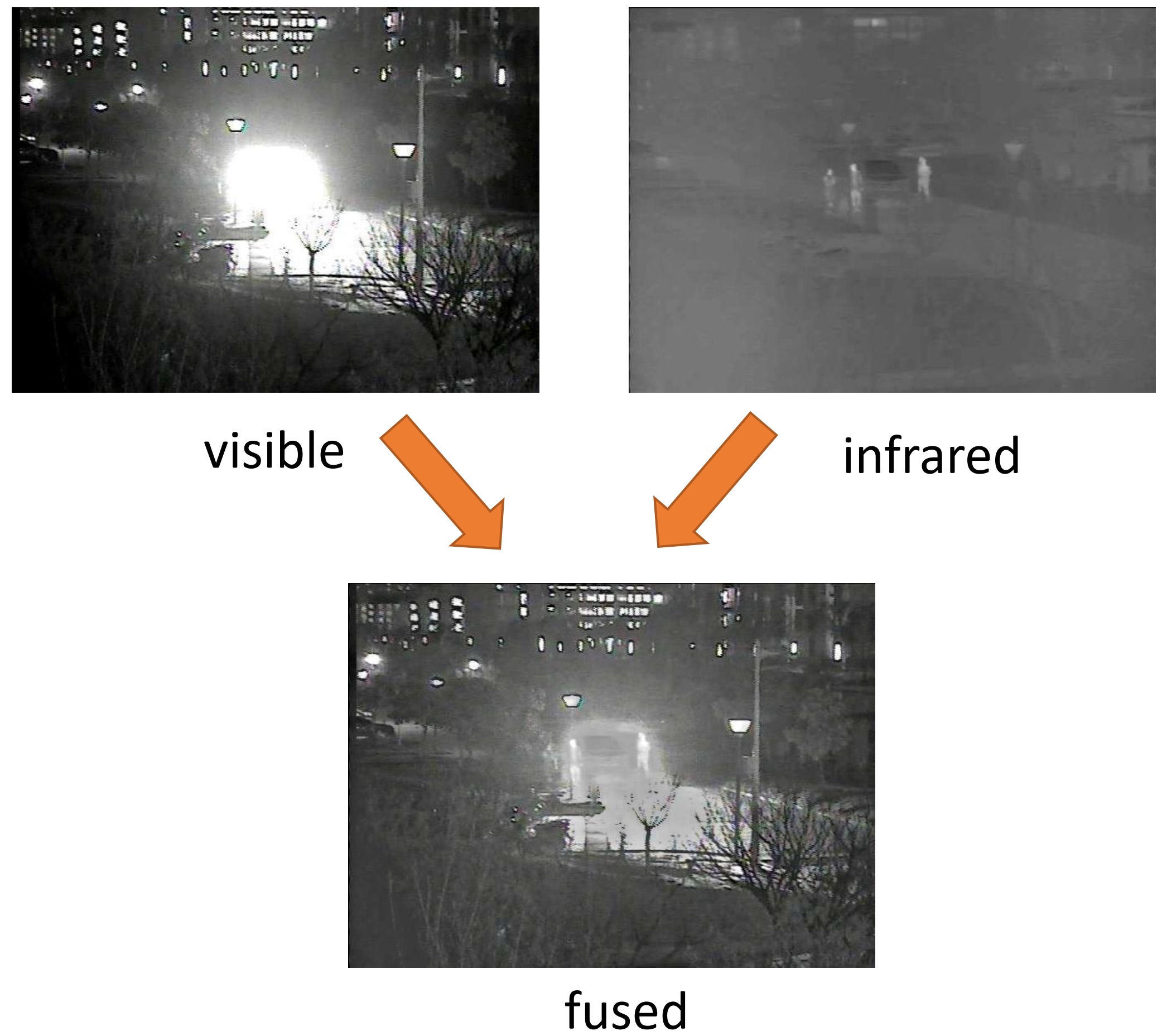}
	\caption{The benefit of visible and infrared image fusion.~The people around the car are invisible in visible image due to  car light.~Although they can be seen in infrared image, the infrared image lacks detail information about the scene.~After fusion, the fused image contains enough details and the people are also visible.}
	\label{fig:example-fusion}
\end{figure}

The target of image fusion is to combine information from different images to generate a single image, which is more informative and can facilitate subsequent processing.~Many image fusion algorithms have been proposed, which can be generally divided into pixel-level, feature-level and decision-level approaches based on the level of fusion.~Also, image fusion can either be performed in the spatial domain or transform domain.~Based on application areas, image fusion technology can be grouped into several types, namely medical image fusion \cite{james2014medical, xia2018novel}, multi-focus image fusion \cite{wang2010multi, liu2017multi, yan2018unsupervised}, remote sensing image fusion \cite{ghassemian2016review}, multi-exposure image fusion \cite{ma2015perceptual, prabhakar2017deepfuse}, visible and infrared image fusion \cite{ma2016infrared, bavirisetti2018new}.~Among these types, the visible and infrared image fusion is one of the most frequently used ones.~This is because that the visible and infrared image fusion can be applied in many applications, for instance object tracking \cite{zhang2019object, zhang2019siamft, li2019rgb, xu2018relative, xu2018object}, object detection \cite{torresan2004advanced, lahmyed2019new, yan2018cognitive}, and biometric recognition \cite{kong2005recent, ariffin2017can}.~Figure \ref{fig:example-fusion} shows an example of visible and infrared image fusion.

\begin{table*}
	\begin{center}
		\caption{Details of some existing visible and infrared image fusion datasets and the proposed dataset.}
		\label{table:dataset}
		\footnotesize
		\begin{tabular}{ccccccc}		
			\hline
			Name         & Image/Video pairs & Image type & Resolution & Year & Results  & Code library\\  \hline
			OSU Color-Thermal Database      & 6 video pairs            & RGB, Infrared  & 320$\times$ 240 & 2005 & No & No\\  
			TNO     &    63   image pairs           & multispectral  &  Various&  2014 & No  &  No\\ 
			VLIRVDIF &      24   video pairs         & RGB, Infrared & 720$\times$480 &  2019 &No  & No\\	
			VIFB    & 21 image pairs   & RGB, Infrared & Various   &  2020  &  Yes  & Yes\\
			\hline
		\end{tabular}
	\end{center}
\end{table*}

\begin{figure*}
	\centering
	\includegraphics[width=16cm]{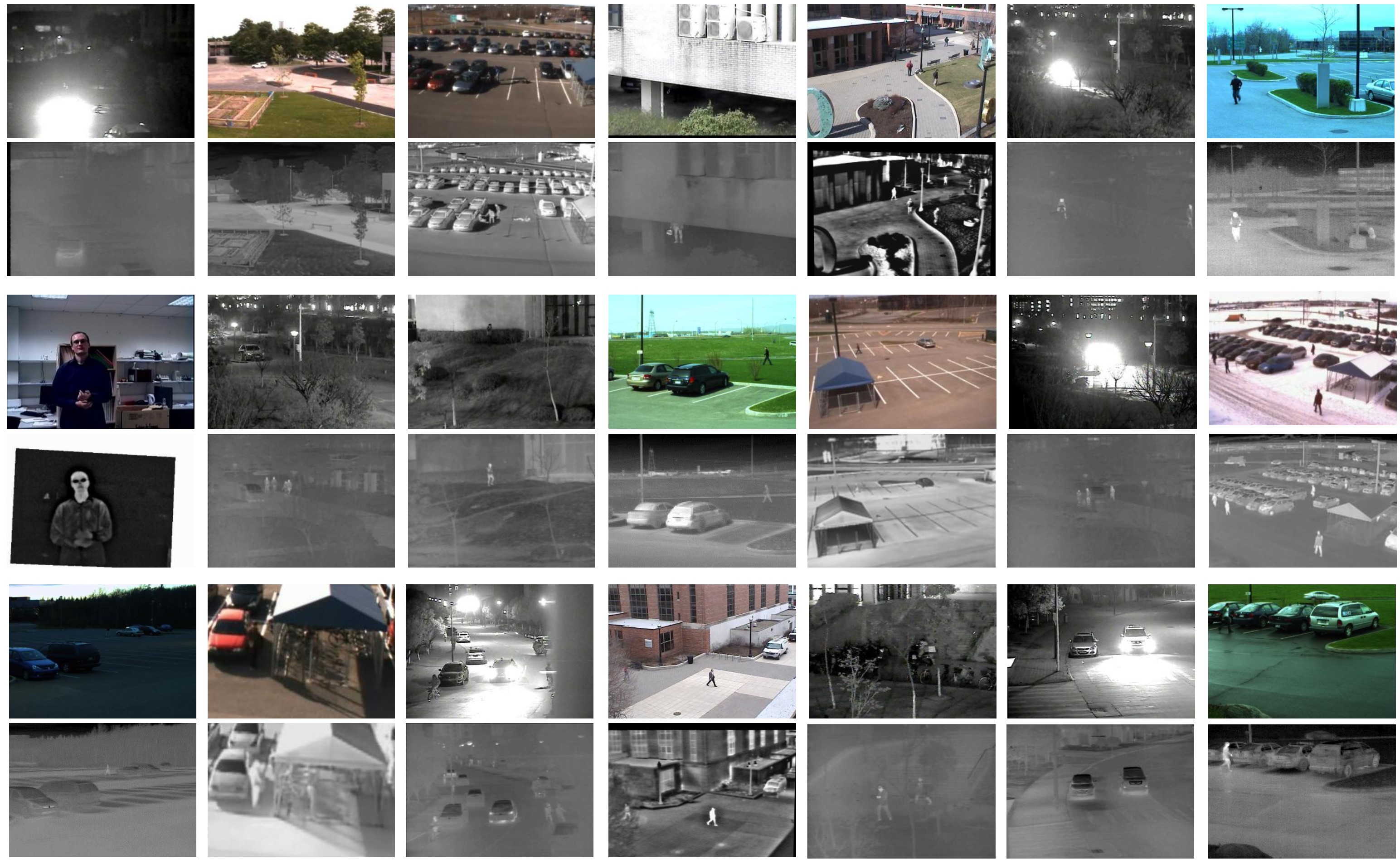}
	\caption{The infrared and visible test set in VIFB.~The dataset includes 21 pairs of infrared and visible images.~The first, third, and fifth row contains RGB images, while the second, fourth, and sixth row presents the corresponding infrared images.}
	\label{fig:example}
\end{figure*}

However, current research on visible and infrared image fusion is suffering from several problems, which hinder the development of this field severely.~First, there is not a well-recognized visible and infrared image fusion dataset which can be used to compare performance under the same standard.~Therefore, it is quite common that different images are utilized in experiments in the literature, which makes it difficult to compare the performance of various algorithms.~Second, it is crucial to evaluate the performance of state-of-the-art fusion algorithms to demonstrate their strength and weakness and to help identify future research directions in this field.~However, although many evaluation metrics have been proposed,  none of them is better than all other metrics.~As a result, researchers normally just choose several metrics which support their methods in the image fusion literature.~This further makes it difficult to objectively compare performances.~Third, although the source codes of some image fusion algorithms have been made publicly available, for example the the DLF \cite{li2018infrareda} and CNN \cite{liu2018infrared}, the interface and usage of most algorithms are different and thus it is inconvenient as well as time-consuming for researchers to perform large scale performance evaluation.

To solve these issues, in this work we build a visible and infrared image fusion benchmark (VIFB) that includes 21 pairs of visible and infrared images, 20 publicly available fusion algorithms and 13 evaluation metrics to facilitate the performance evaluation task \footnote{https://github.com/xingchenzhang/Visible-infrared-image-fusion-benchmark}.

The main contributions of this paper lie in the following aspects:
\begin{itemize}
	\item \textbf{Dataset}.~We created a test set containing 21 pairs of visible and infrared images.~These image pairs are collected from the Internet and several tracking datasets thus covering a wide range of environments and working conditions, such as indoor, outdoor, low illumination, and over-exposure.~Therefore, the dataset is able to test the generalization ability of image fusion algorithms.
	
	\item \textbf{Code library}.~We collected 20 recent image fusion algorithms and integrated them into a code library, which can be easily utilized to run algorithms and compare performance.~Most of these algorithms are published in recent 5 years.~An interface is designed to integrate other image fusion algorithms 
	into VIFB easily.

	\item \textbf{Comprehensive performance evaluation}.~We implemented 13 evaluation metrics in VIFB to comprehensively compare fusion performance.~We have run the collected 20 algorithms on the proposed dataset and performed comprehensive comparison of those algorithms.~All the results are made available for the interested readers to use.
\end{itemize}

\begin{table*}
	\begin{center}
		\caption{Visible and infrared image fusion algorithms that have been integrated in VIFB.}
		\label{table:published}
		\footnotesize
		\begin{tabular}{cccc}		
			\hline
			Method      & Year   & Journal/Conference   & Category   \\ \hline
			ADF \cite{bavirisetti2016fusion}  &  2016    &  IEEE Sensors Journal  &  Multi-scale  \\
			CBF \cite{kumar2015image}   &  2015   &     Signal, image and video processing      &     Multi-scale       \\
			CNN \cite{liu2018infrared}         & 2018    & International Journal of Wavelets, Multiresolution and Information Processing                    & DL-based \\
			DLF \cite{li2018infrareda}    &  2018   &  International Conference on Pattern Recognition         &  DL-based   \\
			FPDE \cite{bavirisetti2017multi}  &  2017   & International Conference on Information Fusion    &   Subspace-based \\	
			GFCE \cite{zhou2016fusion}    &  2016    &   Applied Optics   &     Multi-scale          \\     
			GFF \cite{li2013image}  &  2013   &   IEEE Transactions on Image Processing        &  Multi-scale             \\		
			GTF \cite{ma2016infrared}  &   2016  &  Information Fusion          &       Other       \\      		 			
			HMSD\textunderscore GF\cite{zhou2016fusion}   &  2016 & Applied Optics   &   Multi-scale           \\
			Hybrid\textunderscore MSD \cite{zhou2016perceptual}  & 2016  & Information Fusion  & Multi-scale  \\
			IFEVIP \cite{zhang2017infrared}   & 2017   & Infrared Physics \& Technology   & Other   \\ 	
			LatLRR \cite{li2018infraredc}     &  2018       & 		arXiv						&          Saliency-based          \\ 
			MGFF \cite{bavirisetti2019multi}   &   2019       & Circuits, Systems, and Signal Processing              &  Multi-scale  \\
			MST\textunderscore SR \cite{liu2015general}  & 2015  & Information Fusion & Hybrid \\										 			 	
			MSVD \cite{naidu2011image}   &   2011 & Defense Science Journal     &  Multi-scale  \\	
			NSCT\textunderscore SR \cite{liu2015general}   &2015   & Information Fusion  & Hybrid  \\
			ResNet \cite{li2019infrared}          &  2019   & Infrared Physics \& Technology  &   DL-based  \\
			RP\textunderscore SR \cite{liu2015general}   &  2015 &  Information Fusion & Hybrid   \\
			TIF \cite{bavirisetti2016two}   & 2016      &  Infrared Physics \& Technology  & Saliency-based \\	 
			VSMWLS \cite{ma2017infrared}   &  2017    & Infrared Physics \& Technology &  Hybrid  \\		 
			\hline
		\end{tabular}
	\end{center}
\end{table*}

\section{Related Work}
In this section, we briefly review recent visible and infrared image fusion algorithms.~In addition, we summarize existing visible and infrared image datasets.

\subsection{Visible-infrared fusion methods}
A lot of visible and infrared image fusion methods have been proposed.~Before deep learning is introduced to the image fusion community, 
main image fusion methods can be generally grouped into several categories, namely multi-scale transform-, sparse representation-, subspace-, and saliency-based methods, hybrid models, and other methods according to their corresponding theories \cite{ma2019infrared}.

In the past few years, a number of image fusion methods based on deep learning have emerged \cite{jin2017survey, li2017pixel, liu2018deep, ma2019infrared}.~Deep learning can help to solve several important problems in image fusion.~For example, deep learning can provide better features compared to handcrafted ones.~Besides, deep learning can learn adaptive weights in image fusion, which is crucial in many fusion rules.~Regarding methods, convolutional neural network (CNN) \cite{hermessi2018convolutional, liu2017multi, yan2018unsupervised, xia2018novel, prabhakar2017deepfuse}, generative adversarial networks (GAN) \cite{ma2019fusiongan}, Siamese networks \cite{liu2018infrared}, autoencoder \cite{li2019densefuse} have been explored to conduct image fusion.~Apart from image fusion methods, the image quality assessment, which is critical in image fusion performance evaluation, has also benefited from deep learning \cite{yan2018two}.~It is foreseeable that image fusion technology will develop in the direction of machine learning, and an increasing number of research results will appear in the coming years.

\subsection{Existing dataset}
Although the research on image fusion has begun for many years, there is still not a well-recognized and commonly used dataset in the community of visible and infrared image fusion.~This differs from the visual tracking community where several well-known benchmarks have been proposed and widely utilized, such as OTB \cite{wu2013online, wu2015object} and VOT \cite{VOT_TPAMI}.~Therefore, it is common that different image pairs are utilized in visible and infrared image fusion literature, which makes the objective comparison difficult.

At the moment, there are several existing visible and infrared image fusion datasets, including OSU Color-Thermal Database \cite{davis2007background}\footnote{http://vcipl-okstate.org/pbvs/bench/}, 
TNO Image fusion dataset\footnote{https://figshare.com/articles/TN\textunderscore Image\textunderscore Fusion\textunderscore Dataset/1008029}, and VLIRVDIF \cite{ellmauthaler2019visible}\footnote{http://www02.smt.ufrj.br/~fusion/}.~The main information about these datasets are summarized in Table \ref{table:dataset}.~Actually, apart from OSU, the number of image pairs in TNO and VLIRVDIF is not small.~However, the lack of code library, evaluation metrics as well as results on these datasets make it difficult to gauge the state-of-the-art based on them.

\begin{table*}
	\begin{center}
		\caption{Evaluation metrics implemented in VIFB.~'+' means that a large value indicates a good performance while '-' means that a small value indicates a good performance.}
		\label{table:metrics}
		\footnotesize
		\renewcommand\arraystretch{1.5}
		\begin{tabular}{p{1.6cm}|l|p{4cm}|l|p{1.6cm}|l|p{3cm}|l}	
			\hline
			Category  & Name   & Meaning         & +/- &Category  & Name   & Meaning         & +/-  \\  \hline
			\multirow{4}{1.6cm}{Information theory-based}       & CE \cite{bulanon2009image}  & Cross entropy & - &\multirow{5}{1.6cm}{Image feature-based}    & AG \cite{cui2015detail}  & Average gradient &+ \\	 \cline{2-4} \cline{6-8}			                
			& EN \cite{aardt2008assessment}  & Entropy       & + &  & EI \cite{rajalingam2018hybrid}   & Edge intensity   &+ \\     \cline{2-4}\cline{6-8}
			& MI \cite{qu2002information}  & Mutual information & + & & SD \cite{rao1997fibre} & Standard deviation & + \\  \cline{2-4} \cline{6-8}			& PSNR \cite{jagalingam2015review} & Peak signal-to-noise ration & + &	& SF \cite{eskicioglu1995image}  & Spatial frequency & + \\ 	  \cline{6-8} 					
			&                                  &                             &   &  &  $Q^{AB/F}$ \cite{xydeas2000objective}  & Gradient-based fusion performance& +	\\	 \hline 
			
			\multirow{2}{1.6cm}{Structural similarity-based}    & SSIM \cite{wang2004image}  &  Structural similarity index measure &+  &\multirow{2}{1.6cm}{Human perception inspired}    & $Q_{CB}$ \cite{chen2009new}& Chen-Blum metric   & +    \\	\cline{2-4} \cline{6-8}
			& RMSE \cite{jagalingam2015review}  &  Root mean squared error & - & &  $Q_{CV}$ \cite{chen2007human} & Chen-Varshney metric   &- \\        
			
			\hline
		\end{tabular}
	\end{center}
\end{table*}

\begin{figure*}
	\centering
	\includegraphics[width=17cm]{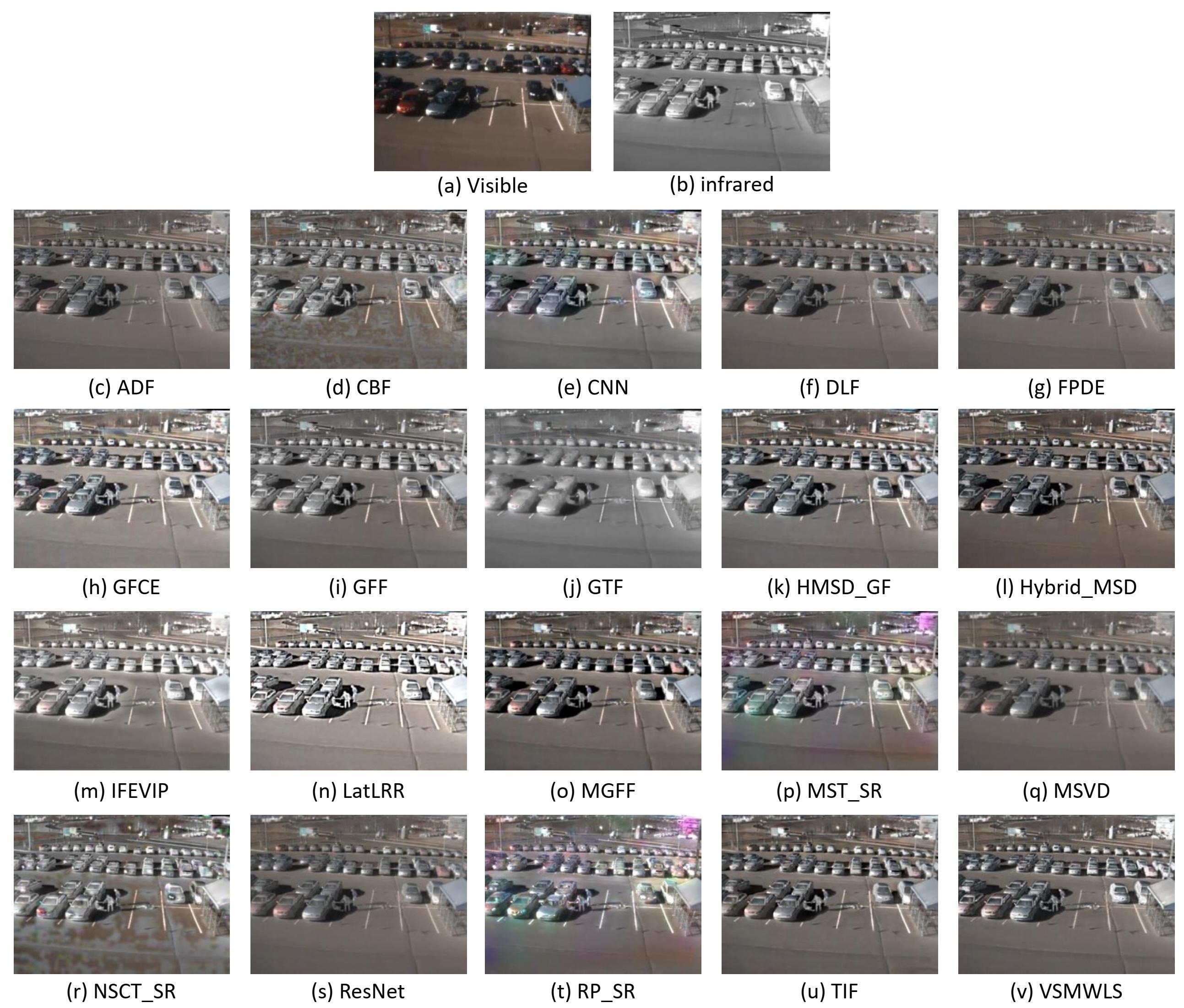}
	\caption{Qualitative comparison of 20 methods on the \textit{fight} image pair shown in Fig.~\ref{fig:example}.}
	\label{fig:qualitative-1}
\end{figure*}

\begin{figure*}
	\centering
	\includegraphics[width=17cm]{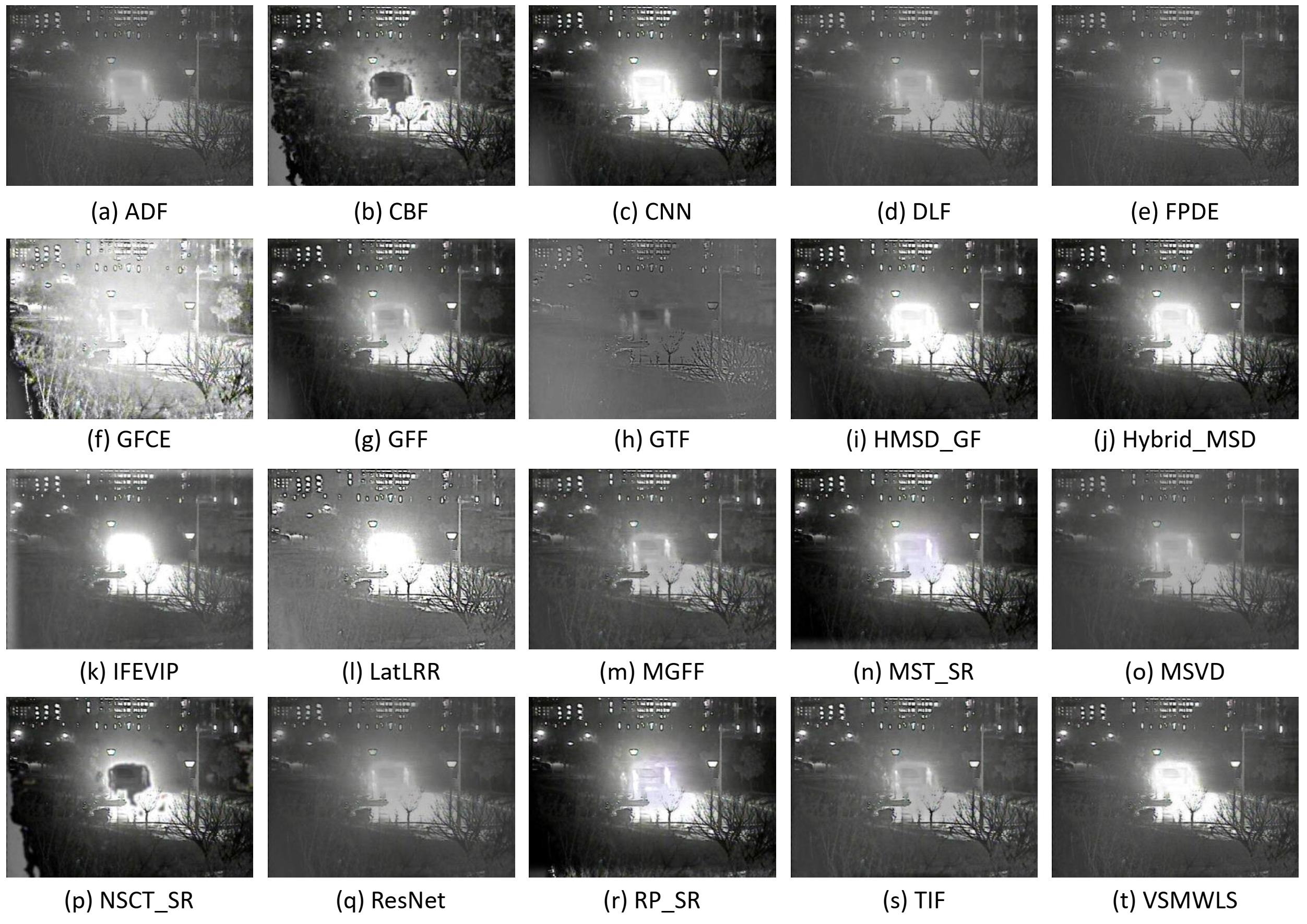}
	\caption{Qualitative comparison of 20 methods on the \textit{manlight} image pair shown in  Fig.~\ref{fig:example-fusion} and Fig.~\ref{fig:example}.}
	\label{fig:qualitative-2}
\end{figure*}

\section{Visible and Infrared Image Fusion Benchmark}
\subsection{Dataset}
The dataset in VIFB, which is a test set, includes 21 pairs of visible and infrared images.~The images are collected by the authors from the Internet\footnote{https://www.ino.ca/en/solutions/video-analytics-dataset/} and fusion tracking dataset \cite{conaire2006comparison, davis2007background, li2019rgb}.~These images cover a wide range of environments and working conditions, such as indoor, outdoor, low illumination, and over-exposure.~Each pair of visible and infrared image has been registered to make sure that the image fusion can be successfully performed.~There are various image resolution in the dataset, such as 320$\times$240, 630$\times$460, 512$\times$184, and 452$\times$332.~Some examples of images in the dataset are given in Fig.~\ref{fig:example}.

\begin{figure*}
	\centering
	\includegraphics[width=15.5cm]{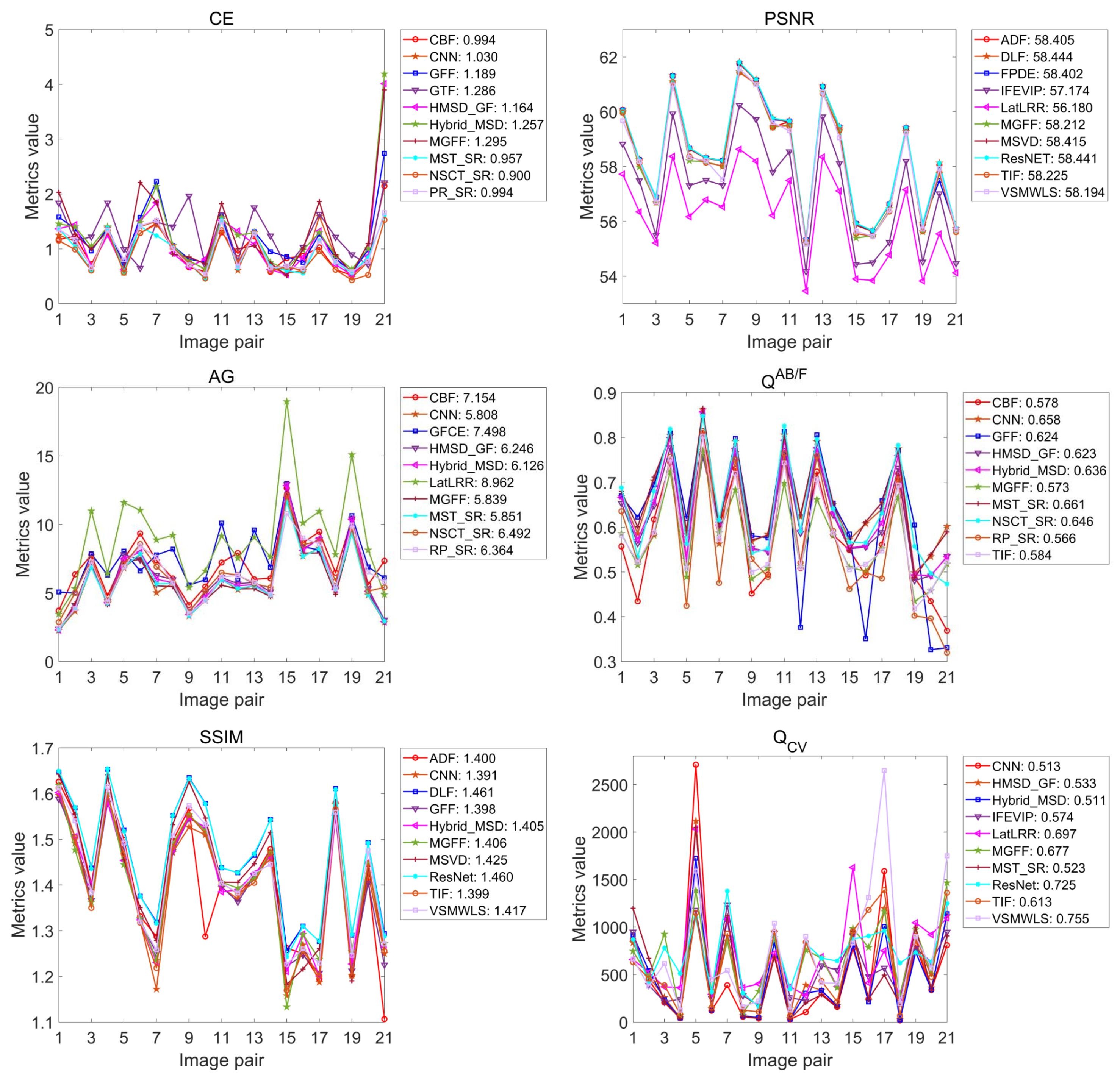}
	\caption{Quantitative comparisons of six metrics of the selected 10 methods on 21 image pairs shown in Fig.~\ref{fig:example}.~The best 10 methods in terms of each evaluation metric are shown.~The values in the legend indicate the average value on 21 image pairs for each method.~From 1 to 21 in the horizontal axis: \textit{carLight, carShadow, carWhite, elecbike, fight, kettle, labMan, man, manCall, manCar, manlight, manWalking, manwithbag, nightCar, peopleshadow, running, snow, tricycle, walking, walking2, walkingnight}.}
	\label{fig:metrics}
\end{figure*}

\subsection{Baseline algorithms}
In recent years, a lot of algorithms have been proposed to perform visible and infrared image fusion.~However, only a part of papers provide the source code.~Besides, these codes have different input and output interfaces, and they may require different running environment.~These factors hinder the usage of these codes to produce results and to perform large-scale performance comparison. 

In VIFB benchmark, we integrated 20 recently published visible-infrared image fusion algorithms including MSVD \cite{naidu2011image}, GFF \cite{li2013image}, MST\textunderscore SR \cite{liu2015general}, RP\textunderscore SR \cite{liu2015general}, NSCT\textunderscore SR \cite{liu2015general}, CBF \cite{kumar2015image}, ADF \cite{bavirisetti2016fusion}, GFCE \cite{zhou2016fusion}, HMSD\textunderscore GF \cite{zhou2016fusion}, Hybrid-MSD \cite{zhou2016perceptual}, TIF \cite{bavirisetti2016two}, GTF \cite{ma2016infrared}, FPDE \cite{bavirisetti2017multi}, IFEVIP \cite{zhang2017infrared}, VSM\textunderscore WLS \cite{ma2017infrared}, DLF \cite{li2018infrareda}, LatLRR \cite{li2018infraredc}, CNN \cite{liu2018infrared}, MGFF \cite{bavirisetti2019multi}, ResNet \cite{li2019infrared}.~Table \ref{table:published} lists more details about these algorithms.~Note that many algorithms were originally designed to fuse grayscale images.~We modified them to fuse color images by fusing every channel of the RGB image with corresponding infrared image.

These algorithms cover almost every kind of visible-infrared fusion algorithms, and most algorithms are proposed in the last five years, which can represent the development of the visible-infrared fusion field to some extent.

To integrate algorithms into VIFB and for the convenience of users, we designed an interface.~By using this interface, other visible-infrared fusion algorithms or their fusion results can be integrated to VIFB to compare their results with those integrated algorithms.

\subsection{Evaluation metrics}
Numerous evaluation metrics for visible-infrared image fusion have been proposed.~As introduced in \cite{liu2012objective}, image fusion metrics can be classified into four types, namely information theory-based, image feature-based, image structural similarity-based, and human perception-based
metrics.~However, none of the proposed metrics is better than all others.~To have comprehensive and objective performance comparison, we implemented 13 evaluation metrics in VIFB.~All evaluation metrics that have been implemented in VIFB and their corresponding categories are listed in Table \ref{table:metrics}.~As can be seen, the implemented metrics in VIFB cover all four categories.~It is convenient to compute all these metrics for each method in VIFB, thus making it easy to compare performances among methods.~Due to the page limits, we leave the detailed introduction to these metrics in the supplementary material.~More information about evaluation metrics can be founded in \cite{liu2012objective, ma2019infrared}.

\section{Experiments}
This section presents experimental results on the VIFB dataset.~Section \ref{subsec:qualitative} and Section \ref{subsec:quantitative} presents qualitative and quantitative performance comparison, respectively.~Section \ref{subsec:runtime} compares the runtime of each algorithm.~All experiments were performed using a computer equipped with an NVIDIA RTX2070 GPU and i7-9750H CPU.~Default parameters reported by the corresponding authors of each algorithm were employed.~Regarding deep learning-based algorithms, the pretrained models provided by their authors were used in this work and we did not retrain those models.~Note that due to the page limits, we just present a part of results here.~More fusion results will be provided in the supplementary materials. 

\begin{table*}
	\begin{center}
		\caption{Average evaluation metric values of all methods on 21 image pairs.~The best three values in each metric are denoted in \xz{red}, \xzg{green} and \xzb{blue}, respectively.~The three numbers after the name of each method denote the number of best value, second best value and third best value, respectively.~Best viewed in color.} 
		\label{table:metrics_average}
		\footnotesize
		\begin{tabular*}{\hsize}{@{}@{\extracolsep{\fill}}cccccccccccccc@{}}     
			\hline
			Method      & AG  & CE  & EI  & EN   &  MI  & PSNR  & $Q^{AB/F}$  & $Q_{CB}$ & $Q_{CV}$ & RMSE   & SF  & SSIM  & SD     \\ \hline
			ADF (0,0,0)  & 4.582 & 1.464 & 46.529  &  6.788  &   1.921 &  58.405 &0.520 &0.474  & 777.8& 0.10426  &14.132 &1.400 & 35.185  \\	
			CBF (0,0,3)  & \xzb{7.154}  & \xzb{0.994} &\xzb{74.590} &7.324&2.161& 57.595&0.578&0.526 &  1575.1 &0.12571&  20.380 &1.171 & 48.544  \\		       
			CNN (1,2,2)  &5.808   & 1.030 & 60.241  &  7.320&\xzb{2.653} &57.932 & \xzg{0.658}&\xzb{0.622} &  \xzg{512.6} &  0.11782 &  18.813 & 1.391 &\xz{60.075} \\
			DLF (3,0,0)  &3.825 &  1.413 &38.569 & 6.724 &2.030  & \xz{58.444} & 0.434&0.445&759.8 &\xz{0.10348} & 12.491 & \xz{1.461} &34.717 \\
			FPDE (0,0,0)  &4.538 & 1.366 & 46.022 &6.766  &1.924 &  58.402 &0.484 & 0.460&  780.1  & 0.10448 &   13.468 & 1.387 &34.931  \\
			GFCE (0,3,0) &\xzg{7.498} &1.931 &\xzg{77.466} &7.266  &1.844&  55.939 & 0.471 &  0.535 &   898.9  & 0.17283&   \xzg{22.463} & 1.134&51.563  \\
			GFF (0,0,0)   &5.326 & 1.189& 55.198  &7.210  &2.638 &  58.100 &0.624 &   0.619 &881.6 & 0.11173& 17.272 & 1.398 &50.059 \\
			GTF (0,0,0)    &4.303  &1.286 &43.664 &6.508 & 1.991 &  57.861 & 0.439 &  0.414&  2138.4  & 0.11772&    14.743& 1.371 &35.130  \\
			HMSD\textunderscore GF (0,1,0) &6.246 &1.164 & 65.034 &7.274 &2.472 &  57.940&  0.623 &  0.604&  533.0  & 0.11600  &19.904&  1.394  &\xzg{57.617} \\	
			Hybrid\textunderscore MSD (1,1,0)&6.126 &1.257 &63.491 &7.304&2.619&  58.173&0.636& \xzg{0.623}&   \xz{510.9} &  0.11020  &  19.659& 1.405 &54.922\\
			IFEVIP (0,0,0) &4.984 &1.339 &51.782 &6.936 &2.248 & 57.174& 0.486 & 0.462 &573.8 &  0.13837 &   15.846 & 1.391 &48.491  \\	 
			LatLRR (3,0,0)  &\xz{8.962} &1.684 &\xz{92.813} &6.909  &1.653 &56.180  & 0.438 &  0.497 &697.3   &0.16862&\xz{29.537} & 1.184 &57.134 \\
			MGFF (0,0,0) &5.839 & 1.295  &60.607&7.114 &1.768 &  58.212 &  0.573 & 0.542 &676.9    & 0.10922&  17.916&  1.406 & 44.290 \\	
			MST\textunderscore SR (2,2,3)   &5.851 & \xzg{0.957} &60.781 &\xzb{7.339}  &\xzg{2.809} & 57.951 &\xz{0.661} &\xz{0.645} &   \xzb{522.7}&  0.11653 &  18.807&1.390 &\xzb{57.314}\\																						
			MSVD (0,0,3)  &3.545 & 1.462 &36.202 & 6.705&1.955 & \xzb{58.415} & 0.332 &0.426& 809.0 & \xzb{0.10417}  &  12.525& \xzb{1.425} &34.372 \\
			NSCT\textunderscore SR (3,0,1) & 6.492  &\xz{0.900} &67.956 &\xz{7.396}   & \xz{2.988}&  57.435 &  \xzb{0.646} & 0.617 &   1447.3&  0.13136 & 19.389 & 1.277&52.475\\
			ResNet (0,3,0)   &3.674  &1.364 &37.255&6.735 &  1.988 & \xzg{58.441} &0.407 & 0.445 &  724.8  &  \xzg{0.10353}&11.736 &\xzg{1.460} &34.940 \\		 			 	
			RP\textunderscore SR (0,1,1)    & 6.364 &0.994 &65.219 &\xzg{7.353}  & 2.336&   57.777 & 0.566& 0.606 & 888.8 &  0.12175 & \xzb{21.171} & 1.332 & 55.808 \\ 
			TIF (0,0,0)        & 5.558 &1.371 &57.839 &7.075 &  1.767 & 58.225 & 0.584 &  0.545 &   613.0 & 0.10875& 17.739 & 1.399 & 42.643 \\
			VSMWLS (0,0,0)   & 5.612 & 1.409 &57.252  & 7.028  &  2.035&  58.194 &0.554  & 0.497 &  754.7 & 0.10921&   17.662  & 1.417  & 46.253 \\	
			\hline
		\end{tabular*}
	\end{center}
\end{table*}

\subsection{Qualitative performance comparison}
\label{subsec:qualitative}
Qualitative evaluation methods are important in fusion quality assessment and they assess the quality of fused images on the basis of the human visual system.~Figure \ref{fig:qualitative-1} presents the qualitative performance comparison of 20 fusion methods on the \textit{fight} image pair.~In this image pair, several people are in the shadow of a car thus can not be seen clearly in the visible image while can be seen in infrared image.~As can be seen, in almost all fused images these people can be seen.~However, the fused images obtained by some algorithms have introduced  artifacts information.~These include CBF, IFEVIP, MST\textunderscore SR, NSCT\textunderscore SR, and RP\textunderscore SR.~Besides, the fused images produced by ADF, CNN, GTF, LatLRR and MSVD do not preserve detail information contained in the visible image well.~Figure \ref{fig:qualitative-1} indicates that the fused images obtained by Hybrid\textunderscore MSD, MGFF, TIF and VSMWLS are more natural for human sensitivity and preserve more details.

Figure \ref{fig:qualitative-2} shows the qualitative comparison of 20 methods on \textit{manlight} image pair.~In this case, the people around the car are invisible in visible images due to over-exposure.~It can be seen that in many fused images, the people around the car are still invisible or not clear, such as those produced by CNN, GFCE, HMSD\textunderscore GF, Hybrid\textunderscore MSD, IFEVIP, LatLRR, and VSMWLS.~Some other fused images have more artifacts which are not presented in original images, such as those obtained by CBF, GFCE, and NSCT\textunderscore SR.~Although the fused images produced by MST\textunderscore SR and RP\textunderscore SR preserve the details in visible image well and the people around the car can be seen clearly, some light purple are introduced in the fused images (near the image center) which are not presented in source images.~The results indicate that GFF and MGFF give better subjective fusion performance for the \textit{manlight} case.

\subsection{Quantitative performance comparison}
\label{subsec:quantitative}
Table \ref{table:metrics_average} presents the average value of 13 evaluation metrics for all methods on 21 image pairs.~As can be seen, the 
NSCT\textunderscore SR obtains the best overall quantitative performance by having 3 best values and 1 third best value.~The LatLRR method and DLF obtain the second best overall performance by having 3 best values.~However, this table indicates clearly that there is not a dominant fusion method that can beat other methods in all or most evaluation metrics.~Besides, from the table one can see that the deep learning-based methods show slightly worse performance than conventional fusion algorithms, although each deep learning-based method performs well in some evaluation metrics.~This is very different from the field of object tracking and detection which is almost dominated by deep learning-based approaches. 


From Table \ref{table:metrics_average} one can also see that the top three algorithms show very different performance in different kinds of metrics.~Specifically, the NSCT\textunderscore SR algorithm obtains the best value in CE, EN and MI, which are all information theory-based evaluation metrics.~The LatLRR algorithm shows the best performance in AG, EI and SF, which are all image feature-based metrics.~The DLF method performs well in RMSE, SSIM and PSNR.~Both RMSE and SSIM are structural similarity-based metrics.~The possible reason is that the authors of these algorithms pay more attention to a specific kind of information when designing these algorithms.~This phenomenon further shows that an image fusion algorithm should be evaluated using various kinds of metrics for a comprehensive comparison, which further indicates the benefits of this study.

Note that although the NSCT\textunderscore SR algorithm obtains the best overall quantitative performance, its qualitative performance is not very good.~As can be seen from Fig.~\ref{fig:qualitative-1} and Fig.~\ref{fig:qualitative-2}, it introduces artifacts in the fused images.~Similarly, the LatLRR also shows good quantitative performance but the qualitative performance is relatively poor.~Specifically, in the \textit{fight} case the LatLRR algorithm loses some details of the visible image while in the \textit{manlight} case it fails to show the target which is invisible in visible images due to over-exposure.~Actually, NSCT\textunderscore SR and LatLRR do not perform very well in $Q_{CB}$ and $Q_{CV}$, which are human perception inspired metrics used to measure the visual performance of the fused image.~The different performance between qualitative and quantitative evaluation clearly shows that both qualitative and quantitative comparison are crucial in image fusion quality evaluation.

To further show quantitative comparison of fusion performances of various methods, the values of six metrics of the 10 selected methods on 21 image pairs are presented in Figure \ref{fig:metrics}.

\subsection{Runtime comparison}
\label{subsec:runtime}
The runtime of algorithms integrated in VIFB is listed in Table \ref{table:runtime}.~As can be seen, the runtime of image fusion methods varies significantly from one to another.~This is also true even for methods in the same category.~For instance, both CBF and GFF are multi-scale methods, but the runtime of CBF is more than 50 times that of GFF.~Besides, multi-scale methods are generally fast and deep learning-based algorithms are slower than others even with the help of GPU.~The fastest deep learning-based method, i.e.~ResNet, takes 4.80 seconds to fuse one image pair.~It should be mentioned that all three deep learning-based algorithms in VIFB do not update the model online, but use pretrained model instead. 

One important application area of visible and infrared image fusion is the RGB-infrared fusion tracking \cite{zhang2019object, zhang2019siamft, zhang2020dsiammft}, where the tracking speed is vital for practical applications.~As pointed out in \cite{zhang2019object}, if an image fusion algorithm is very time-consuming, like LatLRR \cite{li2018infraredc}  and NSCT\textunderscore SR \cite{liu2015general}, then it will not be feasible to develop a real-time fusion tracker based on this image fusion algorithm.~Actually, most image fusion algorithms listed in Table \ref{table:runtime} are computationally expensive in terms of tracking.
\begin{table}
	\begin{center}
		\caption{Runtime of algorithms in VIFB (seconds per image pair)} 
		\label{table:runtime}
		\footnotesize
		\begin{tabular}{ccc}		
			\hline
			Method      & Average runtime & Category \\ \hline
			ADF \cite{bavirisetti2016fusion}  & 1.00 & Multi-scale \\
			CBF \cite{kumar2015image}   &  22.97 & Multi-scale\\
			CNN \cite{liu2018infrared}         &  31.76 & DL-based \\
			DLF \cite{li2018infrareda}    & 18.62 & DL-based \\
			FPDE \cite{bavirisetti2017multi}  &  2.72 & Subspace-based \\	     
			GFCE \cite{zhou2016fusion}    & 2.13  & Multi-scale \\
			GFF \cite{li2013image}  &  0.41 & Multi-scale\\
			GTF \cite{ma2016infrared}  & 6.27  & Other\\      		 	
			HMSD\textunderscore GF\cite{zhou2016fusion}   &  2.76  & Multi-scale   \\
			Hybrid\textunderscore MSD \cite{zhou2016perceptual}  & 9.04 & Multi-scale\\ 
			IFEVIP \cite{zhang2017infrared}   &  0.17 & Other\\ 
			LatLRR \cite{li2018infraredc}     & 271.04 & Saliency-based \\ 
			MGFF \cite{bavirisetti2019multi}   & 1.08 & Multi-scale \\		 			 	
			MST\textunderscore SR \cite{liu2015general}  & 0.76 & Hybrid\\																																								
			MSVD \cite{naidu2011image}   & 1.06 & Multi-scale  \\	
			NSCT\textunderscore SR \cite{liu2015general}   & 94.65  & Hybrid \\ 
			ResNet \cite{li2019infrared}          & 4.80 & DL-based\\
			RP\textunderscore SR \cite{liu2015general}   & 0.86 & Hybrid \\  
			TIF \cite{bavirisetti2016two}   & 0.13 & Saliency-based \\	 
			VSMWLS \cite{ma2017infrared}   &  3.51 & Hybrid \\		 
			\hline
		\end{tabular}
		
	\end{center}
\end{table}

\section{Concluding Remarks}
In this paper, we present a visible and infrared image fusion benchmark (VIFB), which includes a test set of 21 image pairs, a code library consists of 20 algorithms, 13 evaluation metrics and all results.~To the best of our knowledge, this is the first visible and infrared image fusion benchmark to date.~This benchmark facilitates better understanding of the state-of-the-art image fusion approaches, and can provide a platform for gauging new methods. 

We carry out extensive experiments using VIFB to evaluate the performance of all integrated fusion algorithms.~We have several observations 
based on our experimental results.~First, unlike some other fields in computer vision where deep learning is almost dominant, such as object tracking and detection, 
the performances of deep learning-based image fusion algorithms do not show superiority over non-learning algorithms at the moment.~However, due to its strong representation ability, we believe that the deep learning-based image fusion approach will be an important research direction in future.~Second, image fusion algorithms may have different performances in different kinds of evaluation metrics, thus it is necessary to utilize various kinds of metrics to comprehensively evaluate an image fusion algorithm.~Besides, both qualitative and quantitative evaluation are crucial.~Finally, the computational efficiency of visible and infrared image fusion algorithms still need to be improved in order to be applied in real-time applications, such as tracking and detection. 

We will continue extending the dataset and code library of VIFB.~We will also implement more evaluation metrics in VIFB.~We hope that VIFB can serve as a good starting point for researchers who are interested in visible and infrared image fusion.

\textbf{Acknowledgments}.~~This work was sponsored in part by the
National Natural Science Foundation of China under Grant 61973212, in part by the Shanghai Science and Technology Committee Research Project under Grant 17DZ1204304.

{\small
	\bibliographystyle{IEEEtran}
	\bibliography{../../../../../../xingchen}
}

\end{document}